\documentclass{Interspeech2024}

\interspeechcameraready

\usepackage{booktabs}
\usepackage{subcaption}
\usepackage{ragged2e}
\usepackage[dvipsnames]{xcolor}
\usepackage{makecell}
\usepackage{colortbl}
\usepackage{tabularx}
\usepackage{cite}

\title{\textsc{SpeechCaps}: Advancing Instruction-Based Universal Speech Models with Multi-Talker Speaking Style Captioning}

\name[affiliation={1}]{Chien-yu}{Huang}
\name[affiliation={2}]{Min-Han}{Shih}
\name[affiliation={1}]{Ke-Han}{Lu}
\name[affiliation={1}]{Chi-Yuan}{Hsiao}
\name[affiliation={1}]{Hung-yi}{Lee}

\address{
  $^1$National Taiwan University, Taiwan\\
  $^2$University of Southern California, USA}
\email{r08921062@ntu.edu.tw, minhansh@usc.edu, \{d12942024, r12942086, hungyilee\}@ntu.edu.tw}


\begin{document}

\maketitle

\begin{abstract}
Instruction-based speech processing is becoming popular.
Studies show that training with multiple tasks boosts performance, but collecting diverse, large-scale tasks and datasets is expensive.
Thus, it is highly desirable to design a fundamental task that benefits other downstream tasks.
This paper introduces a multi-talker speaking style captioning task to enhance the understanding of speaker and prosodic information.
We used large language models to generate descriptions for multi-talker speech.
Then, we trained our model with pre-training on this captioning task followed by instruction tuning.
Evaluation on Dynamic-SUPERB shows our model outperforming the baseline pre-trained only on single-talker tasks, particularly in speaker and emotion recognition.
Additionally, tests on a multi-talker QA task reveal that current models struggle with attributes such as gender, pitch, and speaking rate.
The code and dataset are available at \url{https://github.com/cyhuang-tw/speechcaps}.
\end{abstract}

\section{Introduction}
The advancement of large language models (LLMs) in natural language processing (NLP) has significantly impacted speech processing research \cite{gong2024listen, wu2023decoder, wang2023slm, nachmani2024spoken, kong2024audio}, particularly in developing instruction-based speech models \cite{gong_ltuas, tang2023salmonn, chu2023qwen, shu2023llasm, hu2024wavllm, lu2024desta}.
Unlike conventional task-specific models trained for fixed tasks, instruction-based models use user prompts to perform various tasks, offering greater flexibility.
A key goal is to achieve emergent capabilities for handling unseen tasks effectively, as done in NLP \cite{wei2021finetuned}.
LTU-AS \cite{gong_ltuas} enhances performance across audio and speech tasks with open-ended question-answering data.
SALMONN \cite{tang2023salmonn} uses ASR and audio captioning, adopting activation tuning to mitigate overfitting.
Qwen-Audio \cite{chu2023qwen} employs a multi-task framework to enhance general audio understanding.
WavLLM \cite{hu2024wavllm} uses curriculum training, starting with elementary tasks and progressing to more complex ones.
DeSTA \cite{lu2024desta} learns speech-text alignment by describing a talker's speaking style before instruction tuning.
These models integrate speech features into LLMs, which are responsible for understanding speech and describing them with natural language.

Dynamic-SUPERB \cite{huang2024dynamic} provides a comprehensive set of 55 tasks designed to assess instruction-based speech models, covering dimensions such as content, semantics, and speaker characteristics.
Surprisingly, existing models perform poorly in speaker and emotion tasks, showing an insufficient understanding of these aspects \footnote{\scriptsize\url{https://github.com/dynamic-superb/dynamic-superb/blob/main/docs/leaderboard.md}}.
This capability is crucial for tasks like speaker verification and emotion recognition, which involve processing multiple talkers or expressive speech, and is essential for developing advanced conversation-related applications.
While multi-task training can improve performance, gathering sufficient data for all task types is often costly or infeasible.
Alternatively, we aim to investigate whether training models on fundamental tasks can enable them to learn general knowledge benefiting several downstream applications.

In this paper, we propose a novel task called multi-talker speaking style captioning as a fundamental task to enhance models' general speech understanding capabilities.
Speaking style captioning uses natural language to describe how a speaker talks, focusing on speaker-specific and prosodic information rather than content.
DeSTA \cite{lu2024desta} is the first model to learn general speech knowledge by pre-training on speaking style captioning tasks, but it only involves single-talker captioning, limiting its potential.
Our proposed task aims to describe each speaker's style, including overlapping talkers, making it more challenging.
Consequently, we created \textsc{SpeechCaps}, the first multi-talker speaking style captioning dataset, synthetically generated from PromptSpeech \cite{guo2023prompttts}.
Then, we developed DeSTA+ using a two-stage pre-training approach that extends from single-talker to multi-talker captioning with \textsc{SpeechCaps}, enhancing its understanding of speaker and prosodic information.
Evaluation results on Dynamic-SUPERB show that DeSTA+ significantly improves performance in speaker and emotion tasks, achieves state-of-the-art results, and demonstrates competitive performance in content and semantic tasks compared to DeSTA, highlighting the effectiveness of the proposed task as a pre-training approach.
Besides, we created a test set in \textsc{SpeechCaps} to directly assess models' capabilities in capturing the speaking styles of different talkers.
The poor performance of baseline models on this test set reflects their weaknesses in speaker and emotion tasks on Dynamic-SUPERB.

\section{Pre-Training Task Generation}
\label{sec:task-generation}

\begin{table*}[t]
    \centering
    \caption{An example of metadata along with its corresponding description and question-answer pairs generated by GPT-4o. Descriptions are generated only for the training set, and question-answer pairs are generated only for the testing set.}
    \label{tab:description_sample}
    \resizebox{.97\linewidth}{!}{
    \begin{tabular}{@{}l@{}}
        \toprule
        \phantomsubcaption\label{tab:description_sample:metadata}(a) \textbf{Metadata} \\
        \midrule
        {\color{OliveGreen}{\textbf{Speaker 1:}}} \{
        gender: female,
        emotion: sad,
        pitch: low,
        speed: slow,
        energy: low,
        start: 0.0,
        end: 3.744
        \}
        \\
        {\color{Orange}{\textbf{Speaker 2:}}} \{
        gender: female,
        emotion: shouting,
        pitch: high,
        speed: medium,
        energy: medium,
        start: 1.176,
        end: 5.106
        \}
        \\
        {\color{BlueViolet}{\textbf{Speaker 3:}}} \{
        gender: male,
        emotion: cheerful,
        pitch: medium,
        speed: fast,
        energy: high,
        start: 5.562,
        end: 9.546
        \}
        \\
        \midrule
        \midrule
        \phantomsubcaption\label{tab:description_sample:description}(b) \textbf{Description (Only in training)} \\
        \midrule
        \parbox[t]{\textwidth}{
        \vspace{-0.75em}
            {\color{OliveGreen}{The audio starts with a female speaker expressing sadness, her voice low and slow, conveying sorrow.}}
            {\color{Orange}{Overlapping slightly, another female voice interjects, shouting with a high pitch and medium speed, her tone filled with intensity.}}
            {\color{BlueViolet}{Following this, a male speaker chimes in, his cheerful and medium-pitched voice moving quickly with high energy, bringing a lively and upbeat atmosphere.}}
        } \\
        \midrule
        \midrule
        \phantomsubcaption\label{tab:description_sample:qa}(c) \textbf{Question-Answer Pairs (Only in testing)} \\
        \midrule
        \color{RoyalBlue}{Q: Among the 3 speakers in the audio, what is the emotion of the speaker who is first in the sequence?} \color{BrickRed}{A: sad}\\
        \color{RoyalBlue}{Q: In this audio, there are 3 speakers. Who, according to their speaking order, speaks at the highest speed?} \color{BrickRed}{A: 3 (the third)} \\
        \bottomrule
    \end{tabular}
    }
\end{table*}

\subsection{Dataset: PromptSpeech}
\label{sec:promptspeech}
\textsc{SpeechCaps} utilizes data from PromptSpeech, which features a variety of expressive utterances from different speakers, specifically designed for training and evaluating PromptTTS, an expressive text-to-speech (TTS) model.
Although the audio files are not publicly available, PromptSpeech provides comprehensive metadata, including transcriptions, biological gender, speaker identity, pitch, speaking rate, energy (loudness), and style prompts.
To generate audio, we used a commercial TTS API\footnote{\scriptsize\url{https://azure.microsoft.com/en-us/products/ai-services/text-to-speech/\#overview}}, which is also used in the official version, to synthesize utterances based on the official metadata.
This process resulted in approximately 50k utterances for the training set, each spoken by a single talker.
We then constructed \textsc{SpeechCaps} from PromptSpeech as described in subsequent sections.

\subsection{multi-talker audio generation}
\label{sec:audio-gen}
For each audio clip, we randomly determined the number of speakers (2 or 3) and sampled one utterance per speaker.
We considered two scenarios for concatenation.
The first scenario involves inserting silence between utterances, with the duration of silence uniformly sampled from the range [0, 1] seconds.
The second scenario involves overlapping utterances, where the overlap duration is uniformly sampled from the range [0.8, 2.4] seconds.
In the overlapping scenario, we simulate a natural conversation where different speakers talk simultaneously.
Finally, we generated 30k audio clips for training data.
The data, including prosodic attributes, starting times, and ending times, were recorded and used to generate text descriptions (Sec.~\ref{sec:dpr-gen}).

\subsection{Description generation}
\label{sec:dpr-gen}

Utilizing language models to generate data and evaluate model output has become a widespread practice in NLP research \cite{kumar2020data, yoo2021gpt}.
However, to the best of our knowledge, no large speech models currently exist that can generate high-quality descriptions of audio.
As an alternative, we used the metadata collected in Sec.~\ref{sec:audio-gen}, formatted it with a crafted prompt, and asked GPT-4o \cite{achiam2023gpt} and Claude3 \cite{claude} to generate a text description for each audio, producing 20k descriptions with GPT-4o and 10k with Claude3.
Table~\ref{tab:description_sample} presents an example of the metadata and its corresponding description.
The description fluently captures each speaker's emotion, pitch, energy, and speaking rate, effectively illustrating their characteristics.

\subsection{Training Overview}
The training framework of DeSTA+ consists of two pre-training stages and one instruction-tuning stage.
In the first stage, \textit{single-talker} speaking style captioning, DeSTA+ learns the basics of speaker and prosodic information.
The second stage, \textit{multi-talker} speaking style captioning, enhances DeSTA+'s ability to identify different talkers and prosodic variations in mixed audios (Sec.~\ref{sec:audio-gen}).
Finally, in the instruction-tuning stage, we use a dataset with various speech tasks and task-specific instructions to develop DeSTA+'s instruction-following capabilities.

\section{Evaluation}
\subsection{Dynamic-SUPERB}
\label{sec:dynamic-superb}
Dynamic-SUPERB offers a variety of speech tasks that necessitate an understanding of diverse speech information, requiring models to leverage their capabilities for various applications, such as speaker verification.
Specifically, Dynamic-SUPERB includes 55 tasks, and each task consists of text instructions, speech, and labels.
A model receives both the text instructions and speech as input and then performs the task based on the instructions.
Based on the specific speech information involved, these tasks are categorized into six dimensions: (1) content, (2) degradation, (3) paralinguistics, (4) semantics, (5) speaker, and (6) audio.
In this paper, we exclude the audio dimension since we focus on enhancing models' understanding of speaker and prosodic information rather than audio understanding.

\subsection{\textsc{SpeechCaps} testing set}
\label{sec:speechcaps-test}
\begin{figure}[htb]
        \centering
        \includegraphics[width=.875\linewidth]{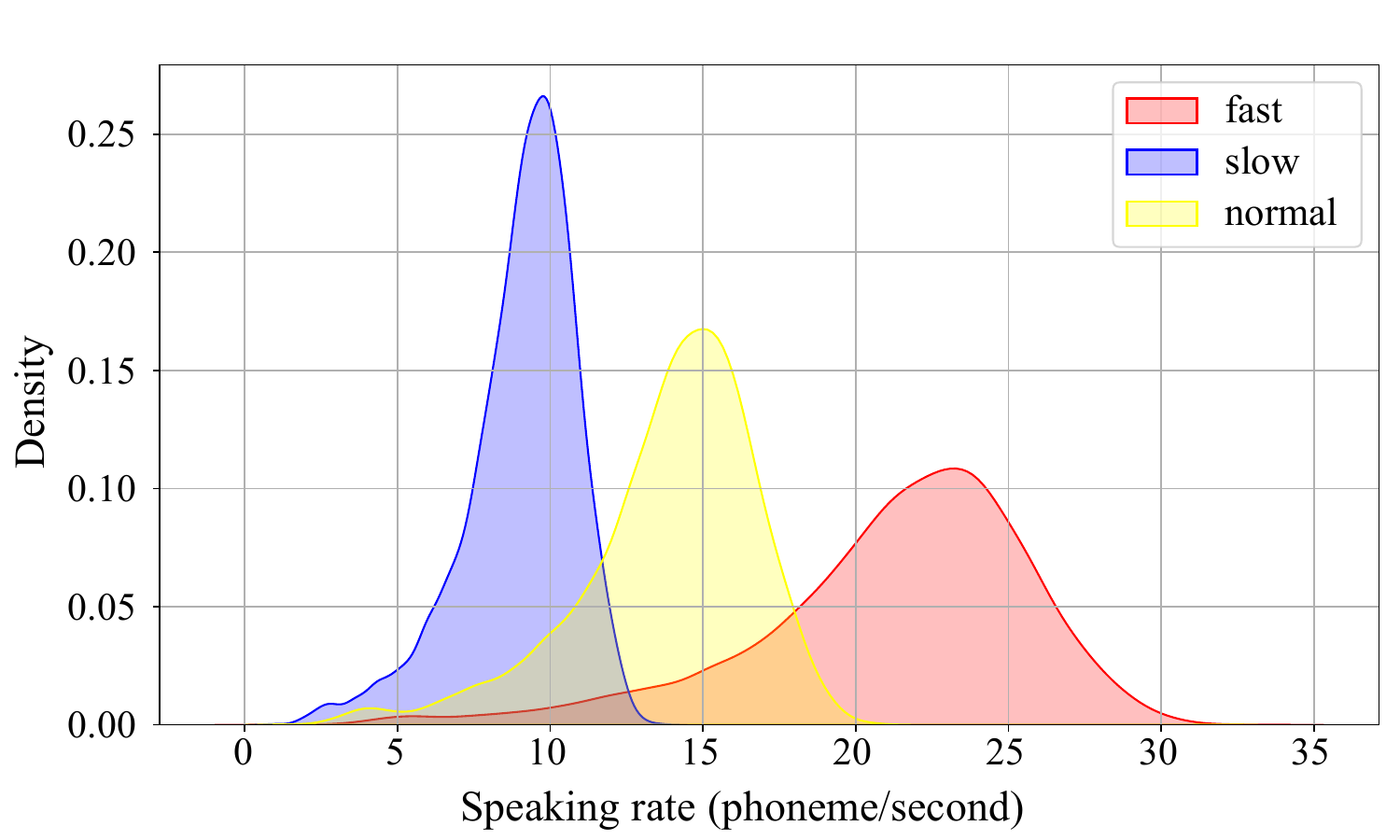}
	\caption{Speaking rate distribution of utterances from a specific speaker (Jenny) in PromptSpeech.}
	\label{fig:ps_org_distribution}
\end{figure}

While Dynamic-SUPERB includes a wide range of tasks, it lacks those that directly assess specific prosodic attributes.
To address this, we created the \textsc{SpeechCaps} test set to evaluate models' understanding of prosodic information.
We built the \textsc{SpeechCaps} test set from the PromptSpeech test set.
Although PromptSpeech provides useful metadata for TTS development, it is not ideal for speech captioning tasks.
Figure~\ref{fig:ps_org_distribution} shows the speaking rate distribution for utterances of a specific speaker in PromptSpeech, with colors indicating the original labels.
We observe overlaps in the distributions of speaking rate levels labeled as low, medium, and high, which can cause ambiguity and complicate evaluation.
Thus, constructing a unified pipeline to relabel the test data is essential.

To relabel the PromptSpeech test set, we used Data-Speech \cite{lacombe-etal-2024-dataspeech} and SoX \cite{sox} to calculate speaking rate (phonemes per second), pitch, and energy.
These distributions are continuous, and we can define two thresholds to divide each attribute into three categories.
However, this also introduces ambiguity for utterances near the thresholds, leading to inconsistent descriptions by different models and complicating evaluation.

To address this, we implemented a filtering process.
For each attribute, we selected data in the lowest 15\%, middle 15\%, and highest 15\%, labeling them as low, medium, and high, respectively.
This filtering was applied to all three attributes: pitch, speaking rate, and energy.
We retained only the utterances falling into these regions across all attributes.
Besides, for pitch, we split utterances by biological gender due to significant differences in pitch distributions between males and females.
After filtering, we got 501 utterances for the test set.

Evaluating speaking style captioning during testing is challenging because a model's output may not include descriptions for all speakers or attributes, and descriptions may be inconsistent across different models.
To reduce ambiguity, we introduced question-answer (QA) pairs as a simplified form to facilitate evaluation.
We devised questions prompting models to identify speakers based on specific attributes. 
Table~\ref{tab:description_sample:qa} shows an example of question-answering for speaking style captioning.
In the question, we ask models to identify the speaker with a particular attribute based on their order or to directly inquire about a specific speaker's emotion.
For pitch-related questions, we specify the speaker's biological gender to mitigate the influence of varying pitch distributions across genders.
Using this approach, we generated 3813 QA pairs for testing.

\subsection{Evaluation metrics}
For each question in the test sets, a model can generate one of three responses: (1) an irrelevant answer, (2) a question-related but incorrect answer, or (3) a correct answer.
The model's ability to follow instructions determines whether it produces irrelevant content (case 1) or a question-relevant answer (cases 2 and 3).
Using the first QA pair in Table~\ref{tab:description_sample:qa} as an example, if a model outputs ``the first speaker,'' it is irrelevant (case 1) because the question asks for emotion, not the speaker's order.
If the response is ``happy,'' it is question-relevant but incorrect (case 2).
While both cases (1) and (2) are incorrect, identifying the reason is crucial.
Therefore, we define the \textbf{instruction-following rate} as the proportion of cases (2) and (3) in the model outputs.
Using LLMs for evaluation has been widely adopted in NLP \cite{wang2023chatgpt, liu2023g, chiang2023can}.
Here, we use GPT-4o, providing it with a prompt that includes the question and the model-generated response.
GPT-4o evaluates whether the output is related to the question.

By calculating the instruction-following rate, we can eliminate irrelevant outputs and focus on evaluating relevant ones.
Exact Match and F1 scores are common metrics in QA but fail to assess semantically similar answers accurately.
Thus, we use GPT-4o to check if a model output aligns with the ground truth.
We design a prompt that includes the question, its ground truth label, and the model output, and GPT-4o then assesses alignment.
We define two types of accuracy: \textbf{overall accuracy}, the percentage of aligned answers in the entire dataset, and \textbf{conditional accuracy}, the percentage of aligned answers among all relevant answers (cases 2 and 3).
In the following sections, accuracy refers to overall accuracy unless specified otherwise.

\section{Experimental Settings}
\label{sec:exp-settings}
\subsection{Models}
We utilized four instruction-based speech models: LTU-AS, SALMONN, Qwen-Audio, and DeSTA.
Each model is primarily constructed with a speech encoder and an LLM and trained with parameter-efficient fine-tuning techniques \cite{hu2021lora, he2021effectiveness}.
The speech features serve as soft prompts that provide various speech information for the LLMs.
LTU-AS and DeSTA extract speech representations and transcriptions from Whisper \cite{radford2023robust}, which are subsequently integrated into LLaMA \cite{touvron2023llama, touvron2023llama2} for further reasoning.
SALMONN adopts a window-level Q-Former \cite{li2023blip} to generate soft embeddings that fuse speech and audio representations from Whisper and BEATs \cite{chen2023beats}.
Qwen-Audio introduces several speech-task-specific tags to encourage knowledge sharing and minimize interference among different tasks.
Based on the size of the LLM, SALMONN is categorized into 7B and 13B versions, and we used the 7B version in the experiments.

\subsection{Implementation details}
\label{sec:impl-details}
For LTU-AS, SALMONN, Qwen-Audio, and DeSTA, we used the official pre-trained models.
To train DeSTA+, we first applied PromptSpeech for single-talker speaking style captioning, followed by \textsc{SpeechCaps} for multi-talker speaking style captioning, and used the Dynamic-SUPERB training set for instruction tuning.
The only difference between DeSTA and DeSTA+ is the multi-talker speaking style captioning in pre-training.
In the multi-talker speaking style captioning stage, we set the learning rate to 1e-4, used a batch size of 12, and trained the model for 5 epochs.
All other hyperparameters matched the official DeSTA implementation.

\section{Results}
\label{sec:results}

\subsection{Dynamic-SUPERB results}

\begin{table}[t]
    \centering
    \caption{Comparison of \textbf{overall accuracy} for various models on the Dynamic-SUPERB dataset, evaluated using GPT-4o.}
    \label{tab:dynamic-superb-results}
    \resizebox{\linewidth}{!}{
    \begin{tabular}{l @{\hspace{2pt}} c @{\hspace{3pt}} c @{\hspace{3pt}} c @{\hspace{3pt}} c @{\hspace{3pt}} c}
    \toprule
    & \multicolumn{5}{c}{\textbf{Model}} \\
    \cmidrule{2-6}
    \textbf{Dimension} & \footnotesize \textbf{LTU-AS} & \footnotesize \textbf{SALMONN} & \footnotesize \textbf{Qwen-Audio} & \footnotesize \textbf{DeSTA} & \footnotesize \textbf{DeSTA+}\\
    \midrule
    (a) Content & 0.445 & 0.521 & 0.622 & 0.913 & 0.877 \\
    (b) Degradation & 0.366 & 0.283 & 0.320 & 0.619 & 0.673 \\
    (c) Paralinguistics & 0.250 & 0.294 & 0.268 & 0.513 & 0.476 \\
    (d) Semantics & 0.362 & 0.509 & 0.480 & 0.743 & 0.729 \\
    (e) Speaker & 0.407 & 0.332 & 0.422 & 0.540 & 0.671 \\
    \bottomrule
    \end{tabular}
    }
\end{table}

We evaluated the proposed pre-training approach's enhancement of model capabilities across a broader range of tasks beyond speech captioning.
Table~\ref{tab:dynamic-superb-results} presents the \textbf{overall accuracy} of the models on Dynamic-SUPERB, using GPT-4o for semantic alignment between model outputs and the ground truth.
Due to space constraints, we do not detail all tasks individually.
Instead, we categorized accuracy results into the following task categories: (a) content, (b) degradation, (c) paralinguistics, (d) semantics, and (e) speaker.
It is important to note that direct comparisons across different dimensions are not feasible due to varying task settings and difficulties.
Alternatively, we compare the performance of different models within the same dimension.
We begin with performance comparisons among LTU-AS, SALMONN, and Qwen-Audio, then shift to DeSTA and DeSTA+. This is because the DeSTA models used an instruction-tuning dataset closely aligned with Dynamic-SUPERB, which could bias comparisons.

From Table~\ref{tab:dynamic-superb-results}, we see that among LTU-AS, SALMONN, and Qwen-Audio, no single model dominated all dimensions, and each had distinct strengths.
Qwen-Audio excelled in content tasks, outperforming LTU-AS and SALMONN, and showed moderate performance in other dimensions except paralinguistics.
LTU-AS, while lagging in content and semantic tasks, was competitive in degradation and speaker tasks.
SALMONN showed superior performance in paralinguistics and semantics but struggled with degradation tasks.
The DeSTA models demonstrated superior performance across all dimensions compared to the above three models, likely due to their instruction-tuning dataset being closely aligned with Dynamic-SUPERB.
Notably, DeSTA+ significantly outperformed in the speaker dimension, indicating an enhanced understanding of speaker information with the proposed approach.
Surprisingly, DeSTA+ also showed improved accuracy in degradation tasks, suggesting that learning from more complex audio scenarios can boost performance across various aspects.
For other dimensions, DeSTA+ performed similarly to DeSTA, indicating that the proposed approach did not degrade performance.

To examine the models' understanding of speaker and prosodic information, we show the performance of speaker verification and emotion recognition in Table~\ref{tab:dynamic-superb-sv-er-results}.
In speaker verification (Table~\ref{tab:ds_sv}), LTU-AS and SALMONN achieved high instruction-following rates (over 85\%) but their accuracies were close to random guesses (50\%).
Qwen-Audio had a lower instruction-following rate (50\%), impacting its overall accuracy on the task.
For emotion recognition (Table~\ref{tab:ds_er}), Qwen-Audio had higher instruction-following rates and accuracy than LTU-AS and SALMONN, but its accuracy was still quite low (around 34\%).
Conversely, though LTU-AS maintained a following rate of around 60\%, its accuracy dropped significantly.
These results show that LTU-AS, SALMONN, and Qwen-Audio have poor capabilities in understanding speakers and emotions.

DeSTA and DeSTA+ achieved much higher instruction-following rates than the other models.
In speaker verification, DeSTA+ outperformed DeSTA by around 10\% accuracy, showing its enhanced ability to distinguish different speakers.
A similar trend was observed in emotion recognition.
DeSTA+ achieved the highest accuracy, demonstrating the effectiveness of the proposed approach.
Although the use of instruction-tuning datasets brought some bias, DeSTA+'s superior performance over DeSTA justifies the effectiveness of the proposed approach, as no new utterances were introduced in training, and we only combined them to create more complex data.

\begin{table}[t]
    \centering
    \caption{Performance comparison of models on speaker verification and emotion recognition tasks in Dynamic-SUPERB.}
    \label{tab:dynamic-superb-sv-er-results}
    \setlength{\tabcolsep}{3pt} 
    \begin{subtable}{\linewidth}
    \caption{speaker verification}
    \label{tab:ds_sv}
        \resizebox{\linewidth}{!}{
        \begin{tabular}{l @{\hspace{2pt}} c @{\hspace{3pt}} c @{\hspace{3pt}} c @{\hspace{3pt}} c @{\hspace{3pt}} c}
        \toprule
        & \multicolumn{5}{c}{\textbf{Model}} \\
        \cmidrule{2-6}
        \textbf{Metric} & \footnotesize \textbf{LTU-AS} & \footnotesize \textbf{SALMONN} & \footnotesize \textbf{Qwen-Audio} & \footnotesize \textbf{DeSTA} & \footnotesize \textbf{DeSTA+}\\
        \midrule
        (a) IF Rate & 0.865 & 0.930 & 0.509 & 0.950 & 0.943 \\
        (b) Overall Acc. & 0.434 & 0.465 & 0.212 & 0.580 & 0.645 \\
        (c) Cond. Acc. & 0.501 & 0.500 & 0.418 & 0.611 & 0.684 \\
        \bottomrule
        \end{tabular}
        }
    \end{subtable}
    \justify
    \begin{subtable}{\linewidth}
    \caption{emotion recognition}
    \label{tab:ds_er}
        \resizebox{\linewidth}{!}{
        \begin{tabular}{l @{\hspace{2pt}} c @{\hspace{3pt}} c @{\hspace{3pt}} c @{\hspace{3pt}} c @{\hspace{3pt}} c}
        \toprule
        & \multicolumn{5}{c}{\textbf{Model}} \\
        \cmidrule{2-6}
        \textbf{Metric} & \footnotesize \textbf{LTU-AS} & \footnotesize \textbf{SALMONN} & \footnotesize \textbf{Qwen-Audio} & \footnotesize \textbf{DeSTA} & \footnotesize \textbf{DeSTA+}\\
        \midrule
        (a) IF Rate & 0.593 & 0.670 & 0.852 & 0.905 & 0.938 \\
        (b) Overall Acc. & 0.055 & 0.200 & 0.341 & 0.598 & 0.658 \\
        (c) Cond. Acc. & 0.094 & 0.299 & 0.400 & 0.660 & 0.701 \\
        \bottomrule
        \end{tabular}
        }
    \end{subtable}
    \justify
    IF: instruction-following, Cond.: conditional.
\end{table}

\subsection{\textsc{SpeechCaps} results}

\begin{table}[t]
    \centering
    \caption{Evaluation of instruction-following rate and accuracy across various models on \textsc{SpeechCaps} test set using GPT-4o.}
    \label{tab:speechcaps_results}
    \setlength{\tabcolsep}{3pt} 
    \resizebox{\linewidth}{!}{
    \begin{tabular}{l c c c c c}
    \toprule
    & \multicolumn{5}{c}{\textbf{Model}} \\
    \cmidrule{2-6}
    \textbf{Metric} & \footnotesize \textbf{LTU-AS} & \footnotesize \textbf{SALMONN} & \footnotesize \textbf{Qwen-Audio} & \footnotesize \textbf{DeSTA} & \footnotesize \textbf{DeSTA+}\\
    \midrule
    (a) IF Rate & 0.379 & 0.519 & 0.783 & 0.607 & 0.483 \\
    (b) Overall Acc. & 0.066 & 0.101 & 0.191 & 0.159 & 0.127 \\
    (c) Cond. Acc. & 0.174 & 0.194 & 0.268 & 0.315 & 0.330 \\
    \bottomrule
    \end{tabular}
    }
\end{table}

In addition to Dynamic-SUPERB, we tested models on the \textsc{SpeechCaps} test set to assess their understanding of prosodic information more directly.
Table~\ref{tab:speechcaps_results} shows the performance of each model.
There were significant differences in instruction-following rates (row (a)).
Qwen-Audio achieved the highest following rate (about 80\%), while LTU-AS lagged significantly behind.
SALMONN and DeSTA had similar rates around 50\%, with DeSTA slightly higher.
Different instruction-following rates directly impacted performance on the test set.
For accurate output, a model must correctly understand the instruction, and a low instruction-following rate leads to lower accuracy.
In row (b), all models showed poor \textbf{overall accuracy}, with Qwen-Audio achieving the highest at about 20\%, and others lagging significantly.
This low performance could result from poor instruction-following or a lack of understanding of speaker and prosodic information.
To this end, we introduced \textbf{conditioned accuracy} in row (c), evaluating only instances where instructions were followed correctly.
Higher conditioned accuracy means a better understanding of speaker and prosodic information.
Although Qwen-Audio had the highest overall accuracy, DeSTA outperformed it in conditioned accuracy by about 5\%.
This suggests DeSTA better understands speaker and prosodic information, likely because it uses the speech captioning task for pre-training.
However, DeSTA's lower instruction-following rate led to its lower overall accuracy.

Last, DeSTA+ had a dropped instruction-following rate than DeSTA but outperformed in conditioned accuracy, and both surpassed the other three models.
This indicates that training with multi-talker data enhanced DeSTA+'s ability to capture speaker and prosodic information, though the different forms of captioning tasks (description in training and QA in testing) impacted its instruction-following capability.

\section{Conclusions}
\label{sec:conclusions}
Instruction-based speech models are gaining popularity across various applications, yet enhancing their fundamental capabilities to benefit several downstream tasks remains unexplored.
This paper introduces a novel task, multi-talker speaking style captioning, as a pre-training approach to enhance model capabilities.
Evaluation on Dynamic-SUPERB shows that this approach significantly improves understanding of speaker and prosodic information, achieving state-of-the-art performance.
Besides, we built a question-answering dataset for prosodic attributes, revealing that the four baseline models cannot capture prosodic information among different talkers.

\bibliographystyle{IEEEtran}
\bibliography{mybib}

\begin{thebibliography}{10}
\providecommand{\url}[1]{#1}
\csname url@samestyle\endcsname
\providecommand{\newblock}{\relax}
\providecommand{\bibinfo}[2]{#2}
\providecommand{\BIBentrySTDinterwordspacing}{\spaceskip=0pt\relax}
\providecommand{\BIBentryALTinterwordstretchfactor}{4}
\providecommand{\BIBentryALTinterwordspacing}{\spaceskip=\fontdimen2\font plus
\BIBentryALTinterwordstretchfactor\fontdimen3\font minus \fontdimen4\font\relax}
\providecommand{\BIBforeignlanguage}[2]{{%
\expandafter\ifx\csname l@#1\endcsname\relax
\typeout{** WARNING: IEEEtran.bst: No hyphenation pattern has been}%
\typeout{** loaded for the language `#1'. Using the pattern for}%
\typeout{** the default language instead.}%
\else
\language=\csname l@#1\endcsname
\fi
#2}}
\providecommand{\BIBdecl}{\relax}
\BIBdecl

\bibitem{gong2024listen}
Y.~Gong, H.~Luo, A.~H. Liu, L.~Karlinsky, and J.~R. Glass, ``Listen, think, and understand,'' in \emph{International Conference on Learning Representations}, 2024.

\bibitem{wu2023decoder}
J.~Wu, Y.~Gaur, Z.~Chen, L.~Zhou, Y.~Zhu, T.~Wang, J.~Li, S.~Liu, B.~Ren, L.~Liu \emph{et~al.}, ``On decoder-only architecture for speech-to-text and large language model integration,'' in \emph{2023 IEEE Automatic Speech Recognition and Understanding Workshop (ASRU)}.\hskip 1em plus 0.5em minus 0.4em\relax IEEE, 2023, pp. 1--8.

\bibitem{wang2023slm}
M.~Wang, W.~Han, I.~Shafran, Z.~Wu, C.-C. Chiu, Y.~Cao, N.~Chen, Y.~Zhang, H.~Soltau, P.~K. Rubenstein \emph{et~al.}, ``Slm: Bridge the thin gap between speech and text foundation models,'' in \emph{2023 IEEE Automatic Speech Recognition and Understanding Workshop (ASRU)}.\hskip 1em plus 0.5em minus 0.4em\relax IEEE, 2023, pp. 1--8.

\bibitem{nachmani2024spoken}
E.~Nachmani, A.~Levkovitch, R.~Hirsch, J.~Salazar, C.~Asawaroengchai, S.~Mariooryad, E.~Rivlin, R.~Skerry-Ryan, and M.~T. Ramanovich, ``Spoken question answering and speech continuation using spectrogram-powered llm,'' in \emph{International Conference on Learning Representations}, 2024.

\bibitem{kong2024audio}
Z.~Kong, A.~Goel, R.~Badlani, W.~Ping, R.~Valle, and B.~Catanzaro, ``Audio flamingo: A novel audio language model with few-shot learning and dialogue abilities,'' in \emph{International Conference on Machine Learning}, 2024.

\bibitem{gong_ltuas}
Y.~Gong, A.~H. Liu, H.~Luo, L.~Karlinsky, and J.~Glass, ``Joint audio and speech understanding,'' in \emph{2023 IEEE Automatic Speech Recognition and Understanding Workshop (ASRU)}, 2023.

\bibitem{tang2023salmonn}
C.~Tang, W.~Yu, G.~Sun, X.~Chen, T.~Tan, W.~Li, L.~Lu, M.~Zejun, and C.~Zhang, ``Salmonn: Towards generic hearing abilities for large language models,'' in \emph{The Twelfth International Conference on Learning Representations}, 2023.

\bibitem{chu2023qwen}
Y.~Chu, J.~Xu, X.~Zhou, Q.~Yang, S.~Zhang, Z.~Yan, C.~Zhou, and J.~Zhou, ``Qwen-audio: Advancing universal audio understanding via unified large-scale audio-language models,'' \emph{arXiv preprint arXiv:2311.07919}, 2023.

\bibitem{shu2023llasm}
Y.~Shu, S.~Dong, G.~Chen, W.~Huang, R.~Zhang, D.~Shi, Q.~Xiang, and Y.~Shi, ``Llasm: Large language and speech model,'' \emph{arXiv preprint arXiv:2308.15930}, 2023.

\bibitem{hu2024wavllm}
S.~Hu, L.~Zhou, S.~Liu, S.~Chen, H.~Hao, J.~Pan, X.~Liu, J.~Li, S.~Sivasankaran, L.~Liu \emph{et~al.}, ``Wavllm: Towards robust and adaptive speech large language model,'' \emph{arXiv preprint arXiv:2404.00656}, 2024.

\bibitem{lu2024desta}
K.-H. Lu, Z.~Chen, S.-W. Fu, H.~Huang, B.~Ginsburg, Y.-C.~F. Wang, and H.~yi~Lee, ``Desta: Enhancing speech language models through descriptive speech-text alignment,'' in \emph{Proc. INTERSPEECH 2024}, 2024.

\bibitem{wei2021finetuned}
J.~Wei, M.~Bosma, V.~Zhao, K.~Guu, A.~W. Yu, B.~Lester, N.~Du, A.~M. Dai, and Q.~V. Le, ``Finetuned language models are zero-shot learners,'' in \emph{International Conference on Learning Representations}, 2021.

\bibitem{huang2024dynamic}
C.-y. Huang, K.-H. Lu, S.-H. Wang, C.-Y. Hsiao, C.-Y. Kuan, H.~Wu, S.~Arora, K.-W. Chang, J.~Shi, Y.~Peng \emph{et~al.}, ``Dynamic-superb: Towards a dynamic, collaborative, and comprehensive instruction-tuning benchmark for speech,'' in \emph{ICASSP 2024-2024 IEEE International Conference on Acoustics, Speech and Signal Processing (ICASSP)}.\hskip 1em plus 0.5em minus 0.4em\relax IEEE, 2024, pp. 12\,136--12\,140.

\bibitem{guo2023prompttts}
Z.~Guo, Y.~Leng, Y.~Wu, S.~Zhao, and X.~Tan, ``Prompttts: Controllable text-to-speech with text descriptions,'' in \emph{ICASSP 2023-2023 IEEE International Conference on Acoustics, Speech and Signal Processing (ICASSP)}.\hskip 1em plus 0.5em minus 0.4em\relax IEEE, 2023, pp. 1--5.

\bibitem{kumar2020data}
\BIBentryALTinterwordspacing
V.~Kumar, A.~Choudhary, and E.~Cho, ``Data augmentation using pre-trained transformer models,'' in \emph{Proceedings of the 2nd Workshop on Life-long Learning for Spoken Language Systems}, W.~M. Campbell, A.~Waibel, D.~Hakkani-Tur, T.~J. Hazen, K.~Kilgour, E.~Cho, V.~Kumar, and H.~Glaude, Eds.\hskip 1em plus 0.5em minus 0.4em\relax Suzhou, China: Association for Computational Linguistics, Dec. 2020, pp. 18--26. [Online]. Available: \url{https://aclanthology.org/2020.lifelongnlp-1.3}
\BIBentrySTDinterwordspacing

\bibitem{yoo2021gpt}
\BIBentryALTinterwordspacing
K.~M. Yoo, D.~Park, J.~Kang, S.-W. Lee, and W.~Park, ``{GPT}3{M}ix: Leveraging large-scale language models for text augmentation,'' in \emph{Findings of the Association for Computational Linguistics: EMNLP 2021}, M.-F. Moens, X.~Huang, L.~Specia, and S.~W.-t. Yih, Eds.\hskip 1em plus 0.5em minus 0.4em\relax Punta Cana, Dominican Republic: Association for Computational Linguistics, Nov. 2021, pp. 2225--2239. [Online]. Available: \url{https://aclanthology.org/2021.findings-emnlp.192}
\BIBentrySTDinterwordspacing

\bibitem{achiam2023gpt}
J.~Achiam, S.~Adler, S.~Agarwal, L.~Ahmad, I.~Akkaya, F.~L. Aleman, D.~Almeida, J.~Altenschmidt, S.~Altman, S.~Anadkat \emph{et~al.}, ``Gpt-4 technical report,'' \emph{arXiv preprint arXiv:2303.08774}, 2023.

\bibitem{claude}
{Anthropic}, ``Claude,'' \url{https://www.anthropic.com}, 2024, large language model.

\bibitem{lacombe-etal-2024-dataspeech}
Y.~Lacombe, V.~Srivastav, and S.~Gandhi, ``Data-speech,'' \url{https://github.com/ylacombe/dataspeech}, 2024.

\bibitem{sox}
\BIBentryALTinterwordspacing
{SoX Contributors}, ``Sox - sound exchange,'' 2024, accessed: 2024-06-16. [Online]. Available: \url{http://sox.sourceforge.net/}
\BIBentrySTDinterwordspacing

\bibitem{wang2023chatgpt}
J.~Wang, Y.~Liang, F.~Meng, Z.~Sun, H.~Shi, Z.~Li, J.~Xu, J.~Qu, and J.~Zhou, ``Is chatgpt a good nlg evaluator? a preliminary study,'' in \emph{Proceedings of EMNLP Workshop}, 2023, p.~1.

\bibitem{liu2023g}
Y.~Liu, D.~Iter, Y.~Xu, S.~Wang, R.~Xu, and C.~Zhu, ``G-eval: Nlg evaluation using gpt-4 with better human alignment,'' in \emph{Proceedings of the 2023 Conference on Empirical Methods in Natural Language Processing}, 2023, pp. 2511--2522.

\bibitem{chiang2023can}
C.-H. Chiang and H.-Y. Lee, ``Can large language models be an alternative to human evaluations?'' in \emph{Proceedings of the 61st Annual Meeting of the Association for Computational Linguistics (Volume 1: Long Papers)}, 2023, pp. 15\,607--15\,631.

\bibitem{hu2021lora}
E.~J. Hu, P.~Wallis, Z.~Allen-Zhu, Y.~Li, S.~Wang, L.~Wang, W.~Chen \emph{et~al.}, ``Lora: Low-rank adaptation of large language models,'' in \emph{International Conference on Learning Representations}, 2021.

\bibitem{he2021effectiveness}
R.~He, L.~Liu, H.~Ye, Q.~Tan, B.~Ding, L.~Cheng, J.~Low, L.~Bing, and L.~Si, ``On the effectiveness of adapter-based tuning for pretrained language model adaptation,'' in \emph{Proceedings of the 59th Annual Meeting of the Association for Computational Linguistics and the 11th International Joint Conference on Natural Language Processing (Volume 1: Long Papers)}, 2021, pp. 2208--2222.

\bibitem{radford2023robust}
A.~Radford, J.~W. Kim, T.~Xu, G.~Brockman, C.~McLeavey, and I.~Sutskever, ``Robust speech recognition via large-scale weak supervision,'' in \emph{International Conference on Machine Learning}.\hskip 1em plus 0.5em minus 0.4em\relax PMLR, 2023, pp. 28\,492--28\,518.

\bibitem{touvron2023llama}
H.~Touvron, T.~Lavril, G.~Izacard, X.~Martinet, M.-A. Lachaux, T.~Lacroix, B.~Rozi{\`e}re, N.~Goyal, E.~Hambro, F.~Azhar \emph{et~al.}, ``Llama: Open and efficient foundation language models,'' \emph{arXiv preprint arXiv:2302.13971}, 2023.

\bibitem{touvron2023llama2}
H.~Touvron, L.~Martin, K.~Stone, P.~Albert, A.~Almahairi, Y.~Babaei, N.~Bashlykov, S.~Batra, P.~Bhargava, S.~Bhosale \emph{et~al.}, ``Llama 2: Open foundation and fine-tuned chat models,'' \emph{arXiv preprint arXiv:2307.09288}, 2023.

\bibitem{li2023blip}
J.~Li, D.~Li, S.~Savarese, and S.~Hoi, ``Blip-2: Bootstrapping language-image pre-training with frozen image encoders and large language models,'' in \emph{International conference on machine learning}.\hskip 1em plus 0.5em minus 0.4em\relax PMLR, 2023, pp. 19\,730--19\,742.

\bibitem{chen2023beats}
S.~Chen, Y.~Wu, C.~Wang, S.~Liu, D.~Tompkins, Z.~Chen, W.~Che, X.~Yu, and F.~Wei, ``Beats: audio pre-training with acoustic tokenizers,'' in \emph{Proceedings of the 40th International Conference on Machine Learning}, 2023, pp. 5178--5193.

\end{thebibliography}

\end{document}